\documentclass{article} 

\usepackage{hyperref}
\usepackage{url}

\usepackage{mathrsfs}
\usepackage{times}
 \usepackage{graphicx}
 \usepackage{psfrag}
 \usepackage{epsfig}
 \usepackage{algorithm}
 \usepackage{algorithmic}
 \usepackage{multicol}
 \usepackage{multirow}
 \usepackage{cite}
 \usepackage{array}
\usepackage{amssymb}
\usepackage[cmex10]{amsmath}
\usepackage{mdwmath}
\usepackage{mdwtab}
\usepackage[tight,footnotesize]{subfigure}
\usepackage{mdwlist}
\usepackage{caption}
\usepackage{slashbox}
\usepackage{yhmath}
\usepackage[amssymb]{SIunits}
\usepackage[usenames]{color}

\newcommand{\p}{\partial}

\newcommand{\M}{\mathcal{M}}
\newcommand{\bx}{\mathbf{x}}
\newcommand{\bb}{\mathbf{b}}
\newcommand{\bn}{\mathbf{n}}
\newcommand{\by}{\mathbf{y}}
\newcommand{\mathd}{\mathrm{d}}

\newcommand{\bu}{\mathbf{u}}
\newcommand{\bg}{\mathbf{g}}
\newcommand{\bh}{\mathbf{h}}

\newcommand{\LP}{\mathcal{L}}
\newcommand{\W}{\mathcal{W}}
\newcommand{\D}{\mathcal{D}}

\newtheorem{theorem}{\textbf{Theorem}}[section]

\title{Harmonic Extension}

\author{
Zuoqiang Shi%
\thanks{Mathematical Sciences Center, Tsinghua University, Beijing, China,
100084. \textit{Email: zqshi@math.tsinghua.edu.cn.}%
} 
\and Jian Sun %
\thanks{Mathematical Sciences Center, Tsinghua University, Beijing, China,
100084. \textit{Email: jsun@math.tsinghua.edu.cn.}%
}
\and
Minghao Tian
\thanks{Department of Mathematical Sciences, Tsinghua University, Beijing, China,
100084. \textit{Email: tianmh0918@hotmail.com.}%
}}

\begin{document}

\maketitle

\begin{abstract}
In this paper, we consider the harmonic extension problem, which is widely
used in many applications of machine learning. We find that the transitional
method of graph Laplacian fails to produce a good approximation of the classical
harmonic function. To tackle this problem, we propose a new method called the 
point integral method (PIM). We consider the harmonic extension problem from the
point of view of solving PDEs on manifolds. 
The basic idea of the PIM method is to approximate the harmonicity using 
an integral equation, which is easy to be discretized from points.
Based on the integral equation, we explain the reason why the transitional
graph Laplacian may fail to approximate the harmonicity in the classical sense
and propose a different approach which we call the volume constraint method (VCM). 
Theoretically, both the PIM and the VCM computes a harmonic function
with convergence guarantees, and practically, they are both simple, which amount
to solve a linear system.  One important application of the harmonic extension in 
machine learning is semi-supervised learning. We run a popular semi-supervised
learning algorithm by Zhu et al.~\cite{ZhuGL03} over a couple of well-known datasets 
and compare the performance of the aforementioned approaches. Our experiments show
the PIM performs the best. 
\end{abstract}

\section{Introduction}
In this paper, we consider the following harmonic extension problem. Let
$X = \{\bx_1, \cdots, \bx_n\}$ be a set of points in $\mathbb{R}^d$ and $B$
be a subset of $X$. Given a function $\bg$ over $B$, let 
$C_\bg = \{ \bu: X\rightarrow \mathbb{R} | \bu_{B} = \bg\}$ be the set of 
functions on $X$ whose restriction to $B$ coincides with $\bg$. Denote $\bu_i = \bu(\bx_i)$. 
The goal of the harmonic extension problem is to find the smoothest function in $C_\bg$. 

A commonly used approach is based on graph Laplacian~\cite{Chung:1997, ZhuGL03}. 
Let $w_{ij}=\exp(-\frac{\|\bx_i-\bx_j\|^2}{4t})$
be the Gaussian weight between $\bx_i, \bx_j$ for some parameter $t$.
Consider the following quadratic energy functional over $C_\bg$. For any $\bu\in C_\bg$
$E(\bu) = \frac{1}{2}\sum_{i, j} w_{ij}(\bu_i-\bu_j)^2.$
The small energy $E(\bu)$ means the function $\bu$ takes similar values at nearby points, 
and the minimizer of this energy is considered as the smoothest function in $C_\bg$. 
It is not difficult to see that the minimizer $\bu$ satisfies that $\LP \bu = 0$ on the points 
in $X\setminus B$ and $\bu_B = \bg$. Here $\LP$ is the weighted graph Laplacian given 
in matrix form as $\LP = \frac{1}{t}(\D - \W)$ where $\W=(w_{ij})$ is the weight matrix and 
$\D = \text{diag}(d_i)$ with $d_i=\sum_j w_{ij}$. We call $\bu$ is discrete harmonic if $\LP \bu = 0$. 
The minimizer $\bu$ can be computed by solving the linear system:
\begin{equation} 
\left\{
\begin{array}{rl}
	\LP(X\setminus B, X) \bu = 0, \\
	\bu(\bx_i)=\bg(\bx_i),& \forall \bx_i \in B
\end{array}
\right.  
\label{eqn:glm} 
\end{equation}
where $\LP(X\setminus B, X)$ is a submatrix of $\LP$ by taking the rows corresponding to the subset $X\setminus B$. 
We call this approach of harmonic extension the graph Laplacian method (GLM). Note that
the factor $\frac{1}{t}$ in $\LP$ is immaterial in GLM but introduced 
to compare with other methods later. 

\begin{figure}[t]
  \centering
  \begin{tabular}{cc}
  \includegraphics[width=0.4\textwidth]{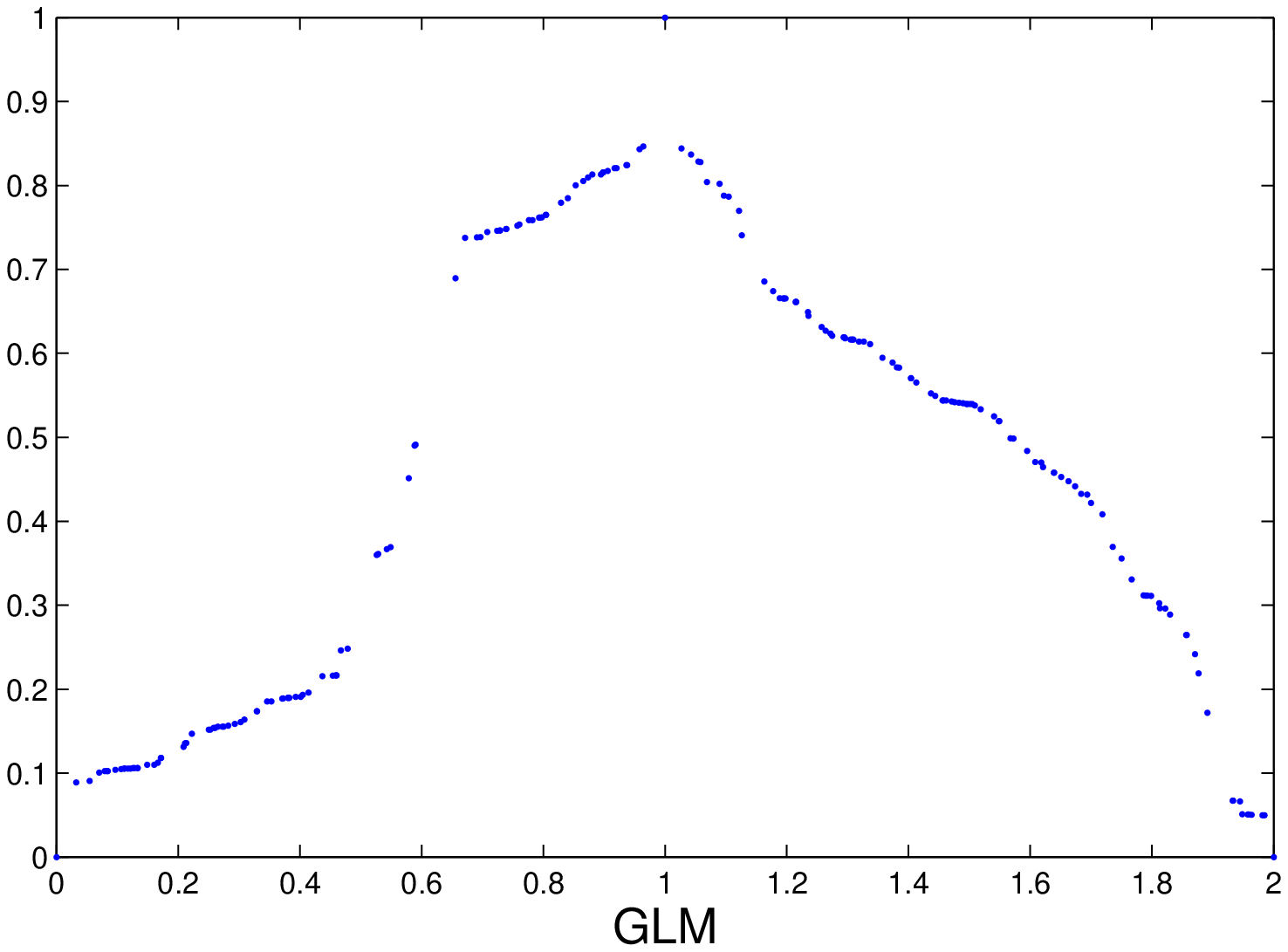} &  \includegraphics[width=0.4\textwidth]{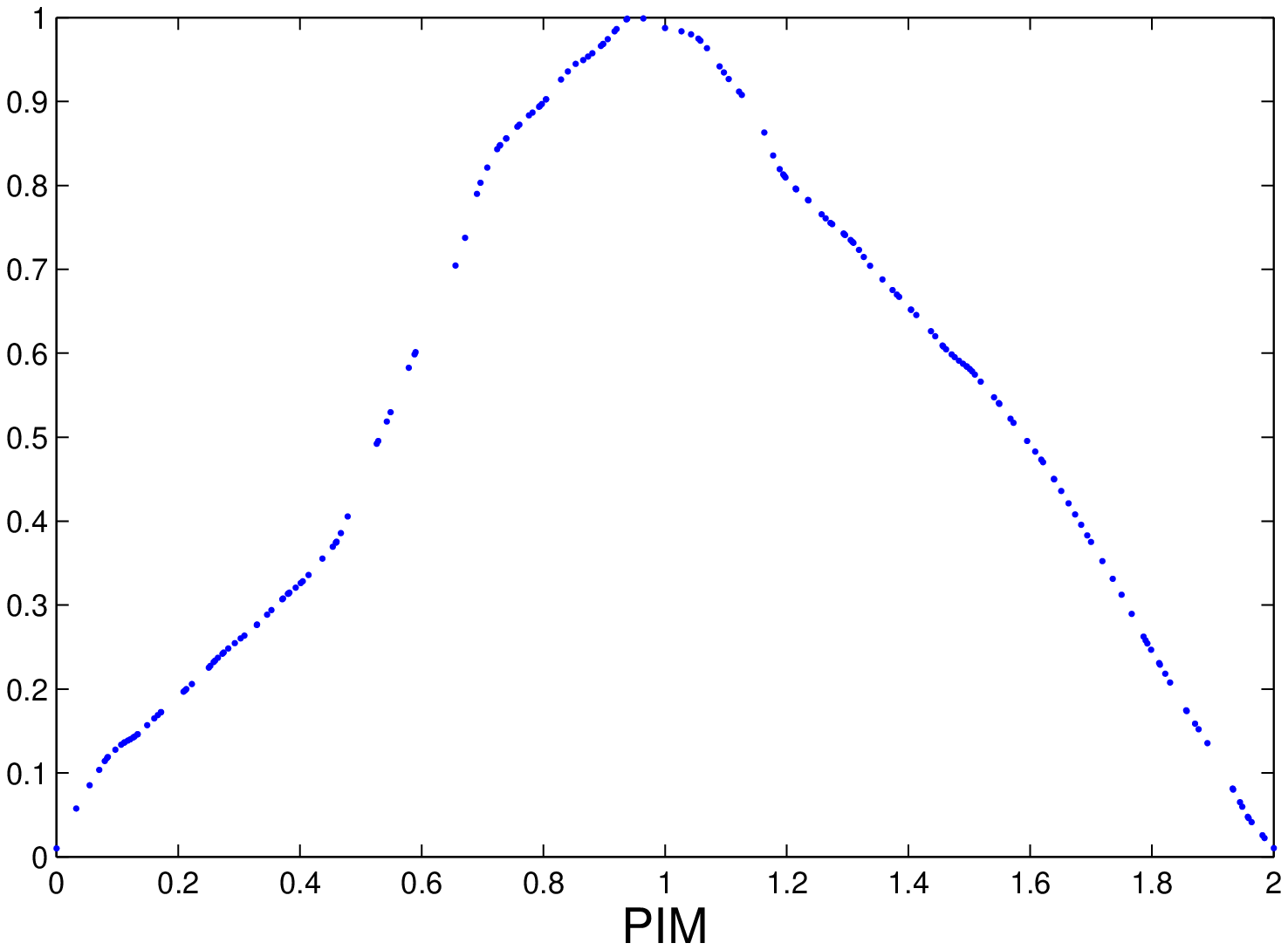} \\
  ~~(a) & ~~(b)\\
  \includegraphics[width=0.4\textwidth]{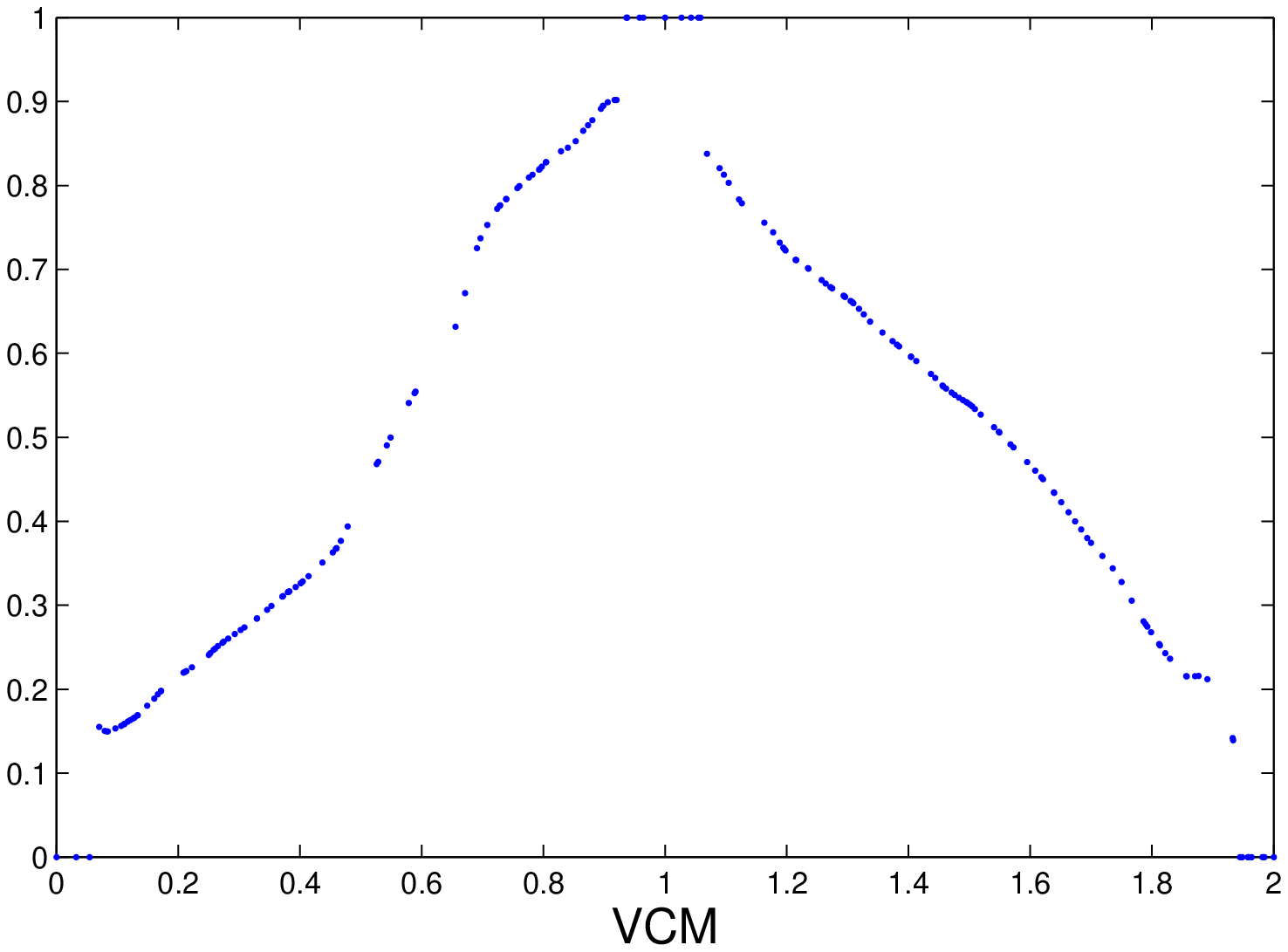} &  \includegraphics[width=0.4\textwidth]{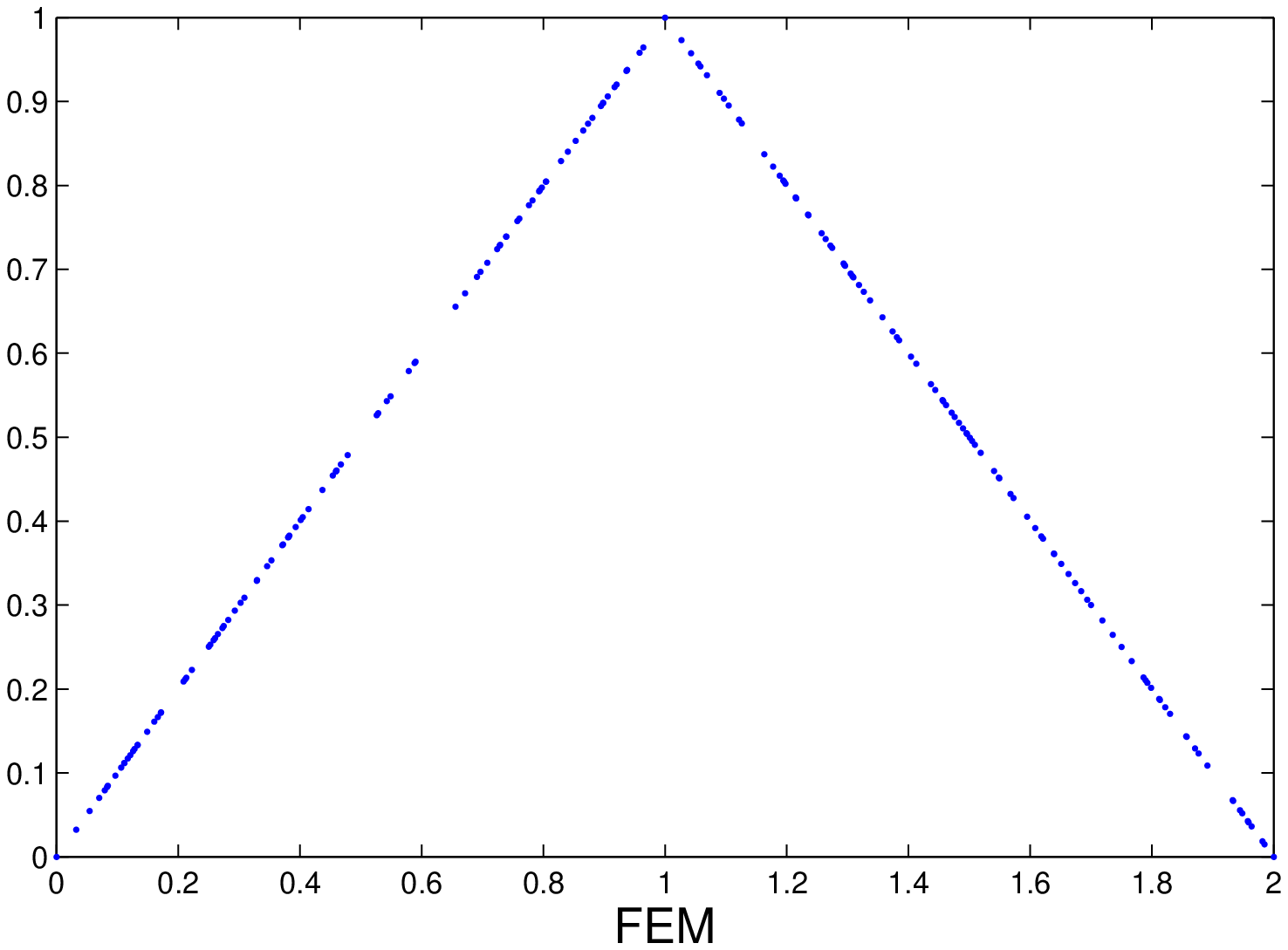} \\
  ~~(c) & ~~(d)
  \end{tabular}
  \caption{One-dimensional examples}
  \label{fig:he}
\end{figure}

Consider the following simple example. Let $X$ be the union of $198$ randomly sampled 
points over the interval $(0, 2)$ and $B= \{0, 1, 2\}$. Set $\bg=0$ at $0, 2$ and 
$\bg=1$ at $1$. We run the above graph Laplacian method over this example. 
Figure~\ref{fig:he} (a) shows the resulting minimizer. It is well-known that the 
harmonic function over the interval $(0, 2)$ with the Dirichlet boundary $\bg$, 
in the classical sense, is a piece linear function, i.e., 
$u(x)=x$ for $x\in (0, 1)$ and $u(x) = 2-x$ for $x\in (1, 2)$;
Clearly, the function computed by GLM does not approximate the harmonic function 
in the classical sense. In particular, the Dirichlet boundary has not been enforced 
properly, and in fact the obtained function is not even continuous near the boundary. 

In this paper, we propose a new method which we call point integral method (PIM) to 
compute harmonic extension. Figure~\ref{fig:he} (b)
shows the harmonic function computed by PIM over the same data, which is a
faithful approximation of the classical harmonic function. The point integral method is very
simple and computes the harmonic extension by solving the following modified linear system:
\begin{equation}
\LP\bu+ \mu \W(X, B) \bu_B = \mu \W(X, B)\bg,
\label{eqn:dirichlet_dis}
\end{equation}
where $\W(X, B)$ is a submatrix of the weight matrix $\W$ by taking the columns corresponding to
the subset $B$.  Here the parameter $\mu$ is a fixed number whose choice will be described 
in Section~\ref{sec:pim}.
We consider the harmonic extension problem from the point of view of solving PDEs on manifolds, 
and derive the point integral method by approximating the Laplace equation using an \emph{integral
equation} which is then discretized using \emph{points}. Note that the Dirichlet boundary may not be
exactly enforced in PIM. Nevertheless, when the points $X$ and $B$ uniform randomly 
sample a submanifold and its boundary respectively in the iid fashion, the harmonic function 
computed by PIM is guaranteed to converge to the one in the classical sense. See Theorem~\ref{thm:error}. 

We will show the derivation of the point integral method, and explain the reason that 
the graph Laplacian method fails to produce a faithful harmonic extension and propose 
two approaches to modify the GLM: one is to thicken the boundary and the other is to 
change weights. Figure~\ref{fig:he} (c) and (d) shows the resulting harmonic functions 
computed by the above approaches. We call the first approach the volume constraint method (VCM). 
The second approach requires an additional structure of meshes which are in general not 
available in machine learning problems. The main purpose that we discuss the second approach
is to show that it is very subtle to choose the weights for the GLM to recover the
classical harmonic extension. 
For the mathematical proof of the convergence of PIM and VCM, the interested readers are 
referred to the papers~\cite{SS14, LSS, S15}. 

One important application of the harmonic extension in machine learning is semi-supervised 
learning~\cite{Zhu:2005}. We will perform the semi-supervised 
learning using the PIM and the VCM over a few well-known data sets, and compare 
their performance to GLM as well as the closely related method by Zhou et al.~\cite{ZhouBLWS03}. 
The experimental results show that the PIM have the best performance and both
the PIM and the VCM out-perform the GLM and the method by Zhou et al.

\subsection{Related work}
The classical harmonic extension problem, also known as the Dirichlet problem for Laplace equation, 
has been studied by mathematicians for more than a century and has many applications 
in mathematics. The discrete harmonicity has also been extensively studied in the graph 
theory~\cite{Chung:1997}. For instance, it is closely related to random walk and electric networks 
on graphs~\cite{Doyle1984}. In machine learning, the discrete harmonic extension and its 
variants have been used for semi-supervised learning~\cite{ZhuGL03, ZhouBLWS03}. 

Much of research has been done on the convergence of the graph Laplacian. 
When there is no boundary, the pointwise convergence of the graph Laplacian to the 
manifold Laplacian was shown in~\cite{BelkinN05, Lafon04diffusion, Hein:2005:GMW, Singer06}, 
and the spectral convergence of the graph Laplacian was shown in~\cite{CLEM_08}. 
When there is boundary, Singer and Wu~\cite{Singer13} and independently Shi and Sun~\cite{SS14} 
have shown that the spectra of the graph Laplacian converge to that of manifold Laplacian 
with Neumann boundary. 

The discrete harmonic extension problem was studied in the community of numerical PDEs and 
its convergence result is well-known if the Laplacian matrix is derived based on finite difference
provided that the points lie on a regular grid, or based on finite element
provided that the points are the vertices of a well-shaped mesh tessellating the domain. 
However, both assumptions on the points are difficult to be satisfied in the applications of 
machine learning.  Du et al.~\cite{DuGLZ12} considered the nonlocal diffusion problems
which is modeled by a similar integral equation. They observed that the regularity of the 
boundary condition can not infer the regularity of the harmonic extension and thus proposed 
to thicken the boundary and employed volume constraint.

\section{Point Integral Method}
\label{sec:pim}
Let $\M$ and $\p\M$ be a submanifold in $\mathbb{R}^d$ and its boundary respectively. 
Given a smooth function $g$ over $\p\M$, the harmonic extension $u$ of $g$ in the classical 
sense is the solution to the following Laplace equation with the Dirichlet boundary: 
\begin{equation} 
\left\{
\begin{array}{rl}
	-\Delta_\mathcal{M} u(\bx)=0,&\bx\in \mathcal{M} \\
	u(\bx)=g(\bx),& \bx\in \p \mathcal{M}
\end{array}
\right.  
\label{eqn:dirichlet} 
\end{equation}
where $\Delta_\mathcal{M}$ is the well-known Laplace-Beltrami operator.\\
We observe that the Laplace equation is closely related to the following integral equation. 
\begin{eqnarray}
\frac{1}{t}\int_{\M} (u(\bx)-u(\by))w_t(\bx, \by) \mathd \by -2\int_{\p \M} \frac{\p u(\by)}{\p \bn} w_t(\bx, \by) \mathd \tau_\by = 0, 
\label{eqn:integral} 
\end{eqnarray}
where $w_t(\bx, \by) = \exp(-\frac{|\bx-\by|^2}{4t})$. 
We will show the derivation of the integral equation later.
Specifically, we proved the following theorem in~\cite{SS14}. 
\begin{theorem}
  \label{thm:local-error}
If $u\in C^3(\M)$ be a harmonic function on $\M$, i.e., $\Delta_\M u =0$,
then we have for any $\bx \in \M$, 
\begin{eqnarray}
\label{eqn:local-error}
\left\|\frac{1}{t}\int_{\M} (u(\bx)-u(\by))w_t(\bx, \by) \mathd \by - 2\int_{\p \M} \frac{\p u(\by)}{\p \bn} w_t(\bx, \by) \mathd \tau_\by \right\|_{L^2(\mathcal{M})}=O(t^{1/4}).
\end{eqnarray}
\end{theorem}
Denote
\begin{eqnarray}
\label{eqn:laplace_op}
L_tu(\bx) = \frac{1}{t}\int_{\M} (u(\bx)-u(\by)) w_t(\bx, \by) \mathd \by, \text{~and}\\
\label{eqn:bdintegral_op}
I_t\frac{\p u}{\p \bn} (\bx) = \int_{\p \M} \frac{\p u(\by)}{\p \bn} w_t(\bx, \by) \mathd \tau_\by.
\end{eqnarray}
Notice that if the point set $X=\{\bx_1, \cdots, \bx_n \}$ samples the submanifold $\M$, then $L_tu$ 
is discretized and well approximated by $\LP \bu$ up to the volume weight $\frac{|\M|}{n}$ where $\bu = (u(\bx_1), \cdots, u(\bx_n))$. 
If we consider the Laplace equation with the Neumann boundary where $\frac{\p u}{\p \bn} = h$ on $\p\M$ is given,   
we can approximate $I_t\frac{\p u}{\p \bn} (\bx)$ using $\sum_{\bb_i\in B} w_t(\bx, \bb_i) h(\bb_i)$
up to the surface area weight $\frac{|\p\M|}{m}$, provided that the point set 
$B=(\bb_1, \cdots, \bb_m) \subset X$ samples $\p\M$.  
Then we can discretize the integral equation~\eqref{eqn:integral} using the linear system
\begin{equation}
\LP \bu - 2\frac{n|\p\M|}{m|\M|}\W(X, B) \bh = 0, 
\end{equation}
where $\bh = (h(\bb_1), \cdots, h(\bb_m))$.
We now use the Robin boundary $u + \beta  \frac{\p u(\by)}{\p \bn} = g$ on $\p\M$ to bridge 
the Dirichlet boundary and the Neumann boundary as follows. 
For a small parameter $\beta$, the above Robin boundary approximates the Dirichlet boundary $u=g$ on $\p\M$. 
At the same time, we can write the Neumann boundary $\frac{\p u}{\p \bn} = \frac{1}{\beta} (g - u_B)$.
Therefore, the harmonic extension problem~\eqref{eqn:dirichlet} can be numerically solved
by the following linear system
\begin{equation}
\LP \bu - \frac{2}{\beta}\frac{n|\p\M|}{m|\M|}\W(X, B) (\bg - \bu_B) = 0
\label{eqn:dirichlet_dis_1}
\end{equation}
which is the same as the linear system~\eqref{eqn:dirichlet_dis} if set $\mu =  \frac{2}{\beta}\frac{n|\p\M|}{m|\M|}$. 
Indeed, we have proved the following theorem in~\cite{SS14} which bounds the difference between 
the harmonic function $u$ solving the classical harmonic extension problem~\eqref{eqn:dirichlet} 
and the harmonic extension $\bu$ using PIM by solving the linear system~\eqref{eqn:dirichlet_dis_1}. 
\begin{theorem}
  \label{thm:error}
Let $u$ be the solution to the problem~\eqref{eqn:dirichlet} and $\bu$ solves
the linear system~\eqref{eqn:dirichlet_dis_1}. Assume $X$ and $B$ sample $\M$ and $\p\M$
uniform randomly in iid fashion respectively. Then there exist sequences of 
 $t(n, m) \rightarrow 0$ and $\beta(t(n, m)) \rightarrow 0 $ so that in probability
\begin{eqnarray}
\label{eqn:error}
\lim_{n, m\rightarrow \infty} \|u - I(\bu)\|_{L^2(\M)} = 0
\end{eqnarray}
where $I(\bu)$ is a function on $\M$ interpolating $\bu$ defined as 
\begin{equation}
\label{eqn:interp_dirichlet}
I(\bu) (\bx) = \frac{ \sum_{\bx_j \in X} w_t(\bx, \bx_j) \bu_j + t\mu\sum_{\bx_j \in B} w_t(\bx, \bx_j) (\bg_j - \bu_j)} {\sum_{\bx_j \in X} w_t(\bx, \bx_j)}.
\end{equation}
\end{theorem}

Although the convergence results hold only when the input point sets sample submanifolds,
the point integral method can apply to any point sets in $\mathbb{R}^d$.
In the examples shown in the paper,  we take
$\beta = 10^{-4}\frac{|\p\M|}{|\M|} $, and thus $\mu = 10^4\frac{n}{m}$. 

\noindent{\bf Integral Equation:~} We now derive the integral equation~\eqref{eqn:integral}. 
Here we assume $\M$ is an open set on $\mathbb{R}^d$. For a general submanifold, the derivation
follows from the same idea but is technically more involved. The interested readers
are referred to~\cite{SS14}. 
Thinking of $w_t(\bx, \by)$ as test functions, by integral by parts, 
we have 
\begin{eqnarray}
\label{eq-integral-part}
\int_\M \Delta u w_t(\bx,\by) \mathd \by
&=& -\int_\M \nabla u \cdot \nabla w_t(\bx,\by) \mathd \by
+\int_{\p \M} \frac{\p u}{\p \bn} w_t(\bx,\by) \mathd \tau_\by
\nonumber\\
&=& \frac{1}{2t}\int_\M (\by-\bx)\cdot\nabla u(\by)w_t(\bx,\by) \mathd \by
+\int_{\p \M} \frac{\p u}{\p \bn} w_t(\bx,\by) \mathd \tau_\by.
\end{eqnarray}
The Taylor expansion of the function $u$ tells us that
\begin{eqnarray}
	u(\by)-u(\bx)=(\by-\bx)\cdot\nabla u(\by)-\frac{1}{2}(\by-\bx)^T\mathbf{H}_u(\by)(\by-\bx)+O(\|\by-\bx\|^3), 
\end{eqnarray}
where $\mathbf{H}_u(\by)$ is the Hessian matrix of u at $\by$. 
Note that $\int_\M \|\by-\bx\|^n w_t(\bx,\by)\mathd \by=O(t^{n/2})$. 
We only need to estimate the following term.
\begin{eqnarray}
  \label{eq-hessian-euler}
  &&\frac{1}{4t}\int_\M (\by-\bx)^T\mathbf{H}_u(\by)(\by-\bx)w_t(\bx,\by) \mathd \by\nonumber\\
  &=&\frac{1}{4t}\int_\M (\by_i-\bx_i)(\by_j-\bx_j)\p_{ij} u(\by)w_t(\bx,\by) \mathd \by\nonumber\\
  &=&-\frac{1}{2}\int_\M (\by_i-\bx_i)\p_{ij} u(\by)\p_j\left({w}_t(\bx,\by)\right) \mathd \by\nonumber\\
  &=&\frac{1}{2}\int_\M \p_j(\by_i-\bx_i) \p_{ij} u(\by){w}_t(\bx,\by) \mathd \by
  +\frac{1}{2}\int_\M (\by_i-\bx_i) \p_{ijj} u(\by){w}_t(\by,\bx) \mathd \by \nonumber\\
  &&-\frac{1}{2}\int_{\p \M}(\by_i-\bx_i) \bn_j \p_{ij} u(\by){w}_t(\bx,\by) \mathd \tau_\by\nonumber\\
  &=& \frac{1}{2}\int_\M \Delta u(\by){w}_t(\bx,\by) \mathd \by
  -\frac{1}{2}\int_{\p \M}(\by_i-\bx_i) \bn_j \p_{ij} u(\by){w}_t(\bx,\by) \mathd \tau_\by+O(t^{1/2})
\end{eqnarray}
Now consider the second summand in the last line is $O(t^{1/2})$. 
Although its $L_\infty(\M)$ norm is of constant order, its $L^2(\M)$ norm
is of the order $O(t^{1/2})$ due to the fast decay of $w_t(\bx, \by)$. 
Therefore, the integral equation~\eqref{eqn:integral} and Theorem~\ref{thm:local-error} follow 
from the equations~\eqref{eq-integral-part}, \eqref{eq-hessian-euler}.


\section{Volume Constraint Method}
From Theorem~\ref{thm:local-error}, we see that the discrete harmonicity $\LP \bu = 0$ is 
not necessary able to approximate the classical harmonicity $\Delta_\M u = 0$, as there is
an extra term $I_t\frac{\p u}{\p \bn}$ of the integration over the boundary $\p\M$ in the 
integral equation~\ref{eqn:integral}. This explains why the graph Laplacian method fails 
to approximate the harmonic extension in the classical sense. In this section, we will 
describe two approaches to modify the graph Laplacian method so that the discrete 
harmonicity $\LP \bu = 0$ approximates the classical harmonicity $\Delta_\M u = 0$. 

The first approach is based on the observation that the Gaussian weight $w_t(\bx, \by)$ 
decays exponentially fast. Let $\M_t = \{\bx\in \M | d_\M(\bx, \p\M) \geq t^{1/2-\delta})\}$ 
for any $\delta > 0$. For any points $\bx \in \M_t$ and $\by \in \p\M$, $w_t(\bx, \by)=o(t^s)$ 
for any $s$ and is very small for small $t$, and so is the term $I_t\frac{\p u}{\p \bn}$. 
Therefore, 
for a point $\bx \in \M_t$, $L_t u(\bx) = 0$ well approximates $\Delta_\M u(\bx) = 0$. 
For the remaining points in the thickened boundary $\M\setminus \M_t$, 
since the harmonic function are smooth, we can approximate $u(\bx) = g(\bar{\bx})$ 
for $\bx \in \M\setminus \M_t$ where $\bar{\bx}$ is the closest point to $\bx$ on the boundary 
$\p\M$. In the discrete setting, let $B_t = \{\bx_i \in X | \bx_i \in \M \setminus \M_t \}$. 
We discretize the harmonic extension problem 
using the following linear system. Denote $\bar{\bx}_i$ the closest point of $\bx_i$ in $B$. 
\begin{equation} 
\left\{
\begin{array}{rl}
	\LP(X\setminus B_t, X) \bu = 0, \\
	\bu(\bx_i)=\bg(\bar{\bx}_i),& \forall \bx_i \in B_t
\end{array}
\right.  
\label{eqn:vcm_dis} 
\end{equation}
This way of enforcing the Dirichlet condition by thickening the boundary 
also appeared in the paper~\cite{DuGLZ12} by Du et al, which they call the volume constraint. 
Du et al. consider the problem of non local diffusion and the nonlocal operator in 
their setting takes the same form as~\eqref{eqn:laplace_op}. 
We call this approach the volume constraint method (VCM).  The following theorem has been proven 
in~\cite{S15} which guarantees the convergence of VCM. 
\begin{theorem}
 \label{thm:vcm_error}
Let $u$ be the solution to the problem~\eqref{eqn:dirichlet} and $\bu$ solves
the linear system~\eqref{eqn:dirichlet_dis_1}. Assume $X$ and $B$ sample $\M$ and $\p\M$
uniform randomly in iid fashion respectively. Then there exists a sequence of 
 $t(n, m) \rightarrow 0$ so that in probability
\begin{eqnarray}
\label{eqn:vcm_error}
\lim_{n, m\rightarrow \infty} \|u - I(\bu)\|_{L^2(\M_t)} = 0
\end{eqnarray}
where $I(\bu)$ is a function on $\M_t$ interpolating $\bu$ defined as 
\begin{equation}
\label{eqn:interp_dirichlet}
I(\bu) (\bx) = \frac{ \sum_{\bx_j \in X} w_t(\bx, \bx_j) \bu_j } {\sum_{\bx_j \in X} w_t(\bx, \bx_j)}.
\end{equation}

\end{theorem}
Similar to the PIM, although the convergence result for the VCM holds only 
when the input point sets sample submanifolds, the VCM can apply to any point sets 
in $\mathbb{R}^d$.

The second approach is to make the term $I_t\frac{\p u}{\p \bn}$ vanish by choosing 
the weights so that $w_t(\bx, \by) = 0$ for any $\by\in \p\M$.
In one-dimensional case, we can use finite element method to obtain 
the weights. Sort the sample points $x_1<\cdots<x_n$. We can set the weight function at $\bx_i$
as the well-known hat function: $w_t(\bx_i, \bx) = \frac{\bx - \bx_{i-1}}{\bx_i - \bx_{i-1}}$  
for $x\in (\bx_{i-1}, \bx_i)$ and $w_t(\bx_i, \bx) = \frac{\bx_{i+1} - \bx}{\bx_{i+1} - \bx_{i}}$ 
for $\bx\in (\bx_{i}, \bx_{i+1})$, which vanishes at the boundary. 
Figure~\ref{fig:he}(d) shows the resulting harmonic extension computed by solving 
the linear system~\eqref{eqn:glm}. 
In higher-dimensional case, if we are given a triangular mesh of the submanifold having
the sample points $X$ as vertices, we can again use finite element method to obtain the 
weights. We omit the detailed derivation of the weights in higher dimensional 
case. From the standard analysis of finite element method, with such weights, 
the solution computed by the linear system~\ref{eqn:glm} of GLM approximates the classical
harmonic extension~\eqref{eqn:dirichlet} with convergence guarantees. However, 
in the typical setting of machine learning or data analysis, this type of weights 
is very difficult, if not impossible, to obtain. Therefore, although 
it is popular in scientific computing, this approach is in general not applicable 
to machine learning problems.

\section{Semi-supervised Learning}
In this section, we briefly describe the algorithm of semi-supervised learning
based on the harmonic extension proposed by Zhu et al.~\cite{ZhuGL03}. We plug into
the algorithm the aforementioned approaches of harmonic extension, 
and apply them to several well-known data sets, and compare their performance. 

Assume we are given a point set $X=\{\bx_1 ,\cdots, \bx_m, \bx_{m+1} ,\cdots, \bx_n \}\subset \mathbb{R}^d$, 
and a label set $\{1,2,\cdots,l\}$, and the label assignment on the first $m$ points 
$L: \{\bx_1, \cdots, \bx_m\} \rightarrow \{1,2,\cdots,l\}$.  
In a typical setting, $m$ is much smaller than $n$. The purpose of the semi-supervised learning 
is to extend the label assignment $L$ to the entire $X$, namely, infer the labels for the unlabeled points.

Think of the label points as the boundary $B=\{\bx_1, \cdots, \bx_m \}$. For the label $i$, 
we set up the Dirichlet boundary $\bg^i$ as follows. If a point $\bx_j \in B$ is labelled as 
$i$, set $\bg^i(\bx_j) = 1$, and otherwise set $\bg^i(\bx_j) = 0$. Then we compute the 
harmonic extension $\bu^i$ of $\bg^i$ using the aforementioned approaches. In this way, 
we obtain a set of $l$ harmonic functions $\bu^1, \bu^2, \cdots, \bu^l$. We label $\bx_j$
using $k$ where $k=\arg\max_{i\le l} \bu^i(\bx_j)$. The algorithm is summarized in Algorithm~\ref{alg:ssl}.  
Note that this algorithm is slightly different from  the original algorithm by Zhu et al.~\cite{ZhuGL03} 
where only one harmonic extension was computed by setting $\bg^i(\bx_j) = k$ if $\bx_j$ has a label $k$.

\begin{algorithm*}[h]
\floatname{algorithm}{Algorithm}
\caption{Semi-Supervised Learning}
\label{alg:ssl}
\begin{algorithmic}[1]
\REQUIRE A point set $X=\{\bx_1 ,\cdots, \bx_m, \bx_{m+1} ,\cdots, \bx_n \}\subset \mathbb{R}^d$ and 
a partial label assignment $L: \{\bx_1, \cdots, \bx_m\} \rightarrow \{1,2,\cdots,l\}$
\ENSURE  A complete label assignment $L: X \rightarrow \{1,2,\cdots,l\}$
\FOR {$\;i=1:l\quad$}
\FOR {$\;j=1:m\quad$} 
\STATE Set $\bg^i(\bx_j) = 1$ if $L(\bx_j) = i$, and otherwise set $\bg^i(\bx_j) = 0$.
\ENDFOR
\STATE Compute the harmonic extension $\bu^i$ of $\bg^i$.
\ENDFOR
\FOR {$\; j=m+1:n $} 
\STATE  $L(\bx_j)=k$ where  $k=\arg\max_{i\le l} \bu^i(\bx_j)$. 
\ENDFOR
\end{algorithmic}
\end{algorithm*}

\subsection{Experiments}
We now apply the above semi-supervised learning algorithm to a couple of well-known data sets: MNIST and 20 Newsgroups. 
We do not claim the state of the art performance on these datasets. The purpose of these experiments is to compare the 
performance of different approaches of harmonic extension. We also compare to the closely related method of local and 
global consistency by Zhou et al.~\cite{ZhouBLWS03}. 

\noindent{\bf MNIST~: }
In this experiment, we use the MNIST of dataset of handwritten digits~\cite{mnist}, 
which contains $60k$  $28 \times 28$ gray scale digit images with labels. 
We view digits $0\sim 9$ as ten classes. Each digit can be seen as a point in
a common 784-dimensional Euclidean space. 
We randomly choose 16k images.  Specifically, there are 1606, 1808,
1555, 1663, 1552, 1416, 1590, 1692, 1521 and 1597 digits in $0 \sim 9$
class respectively. 

To set the parameter $t$, we build a graph by connecting a point $x_i$ to 
its $10$ nearest neighbors under the standard Euclidean distance. 
We compute the average of the distances for $x_i$ 
to its neighbors on the graph, denoted $h_{i}$. Let $h$ be the average
of $h_i$'s over all points and set $t= h^2$. The distance $|\bx_i-\bx_j|$
is computed as the graph distance between $x_i$ and $x_j$.  
In the method of local and global consistency, we follow the paper~\cite{ZhouBLWS03}
and set the width of the RBF kernel to be $0.3$ and the parameter
$\alpha$ in the iteration process to be $0.3$.

\begin{figure}
  \centering
  \begin{tabular}{cc}
  \includegraphics[width=0.45\textwidth]{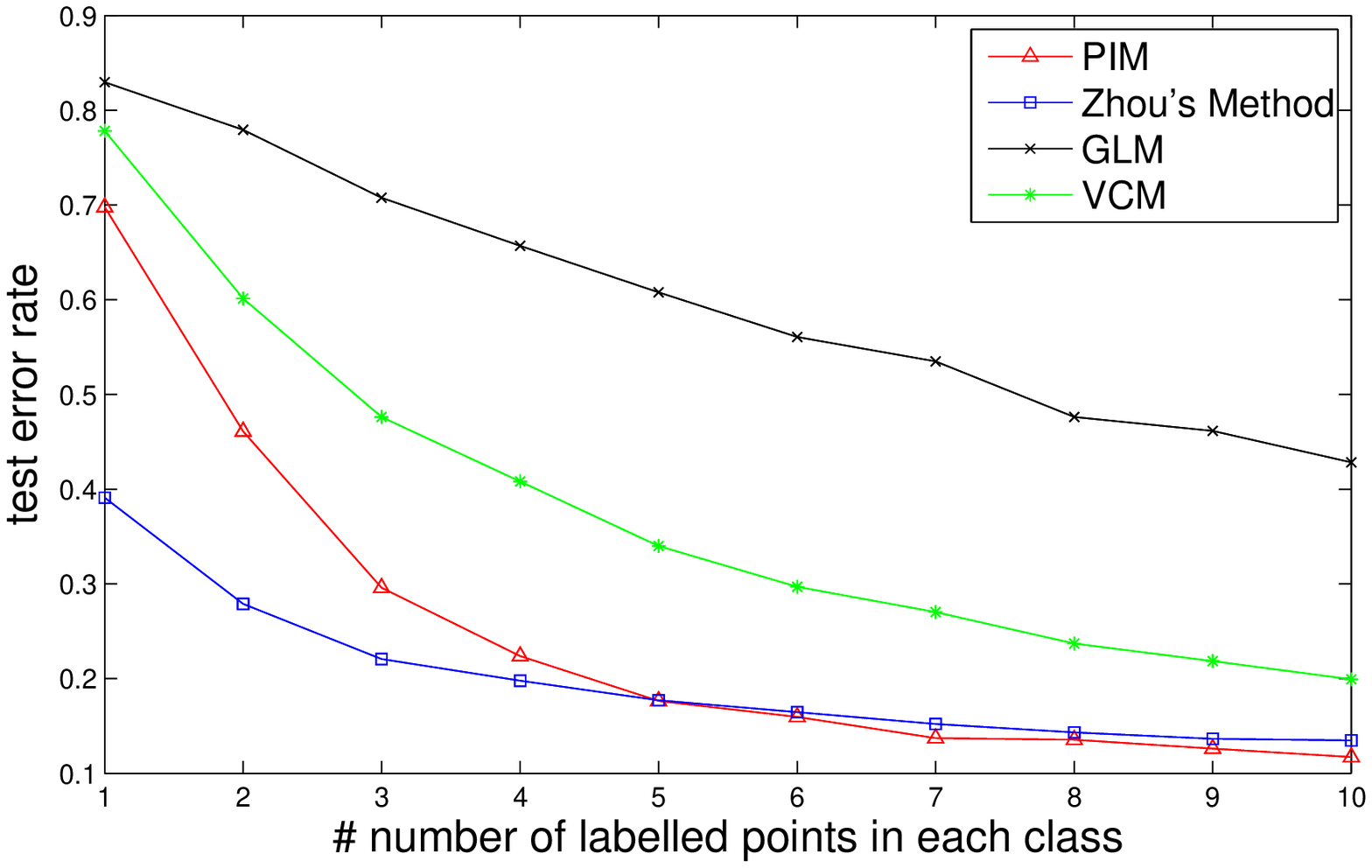} &  \includegraphics[width=0.45\textwidth]{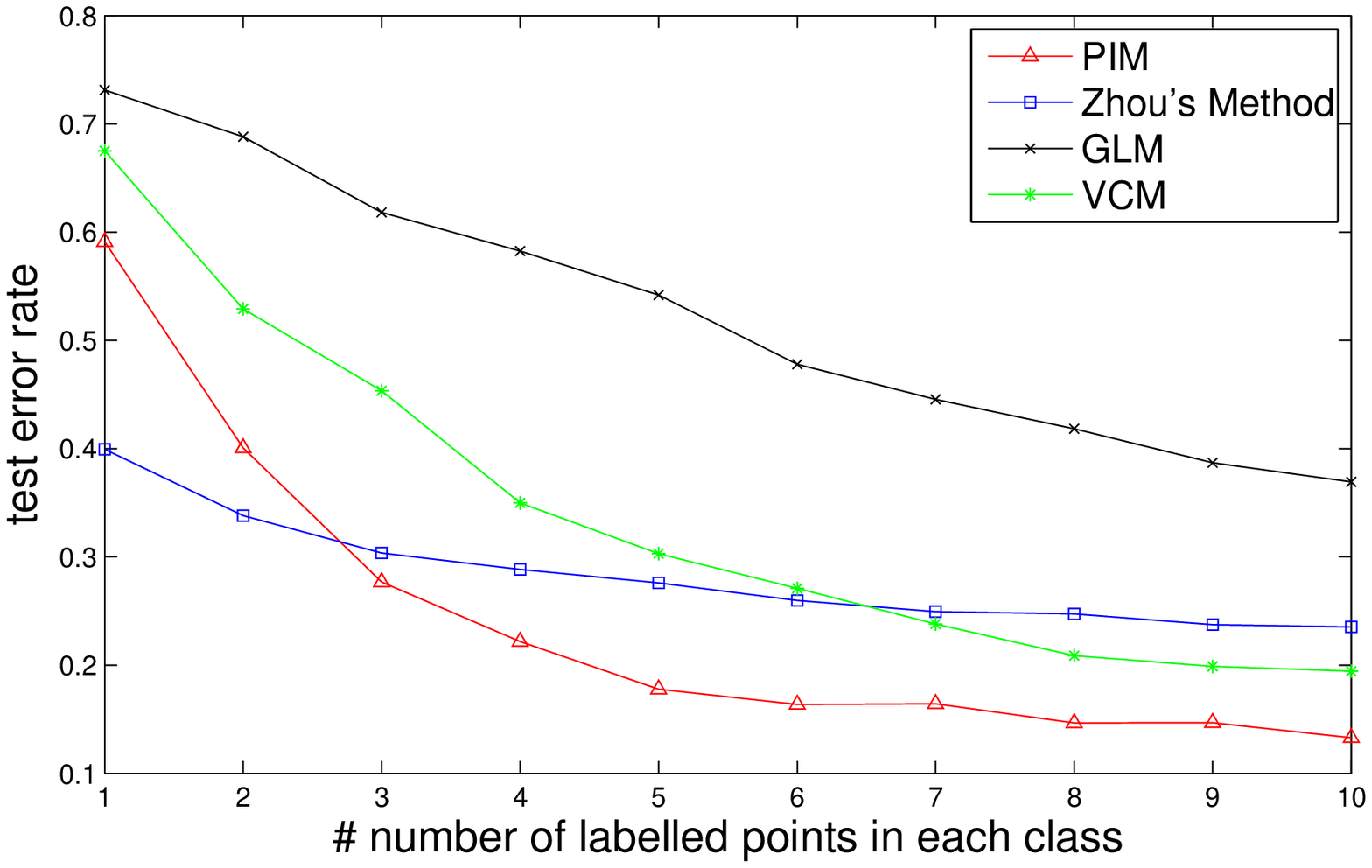} \\
  (a) & (b)\\
  \end{tabular}
  \caption{(a) the error rates of digit recognition with a 16000-size subset of MNIST dataset; (b) the error rates of text classification with 20-newsgroups.rec(a 8014-dimensional space with 3970 data points).}
  \label{fig:er}
\end{figure}

For a particular trial, we choose $k$ $(k=1,2,\cdots,10)$ images randomly from
each class to assemble the labelled set $B$ and assume all the other images
are unlabelled. For each fixed $k$, we do $100$ trials. The error bar of the 
tests is presented in Figure~\ref{fig:er}~(a). It is quite clear that the PIM 
has the best performance when there are more than 5 labelled points in each class, 
and the GLM has the worst performance. 


\noindent{\bf Newsgroup:~}
In this experiment, we use the 20-newsgroups dataset, which is a classic dataset 
in text classification. We only choose the articles from topic \textit{rec} 
containing four classes from the version 20-news-18828. We use Rainbow (version:20020213) 
to pre-process the dataset and finally vectorize them. The following command-line 
options are required\footnote{all the following options are offered by Rainbow}:
(1)\textit{-\,-skip-header}: to avoid lexing headers; (2)\textit{-\,-use-stemming}:
to modify lexed words with the `Porter' stemmer; (3)\textit{-\,-use-stoplist}: to toss
lexed words that appear in the SMART stoplist;
(4)\textit{-\,-prune-vocab-by-doc-count=5}: to remove words that occur in 5 or
fewer documents; Then, we use TF-IDF algorithm to normalize the word count
matrix. Finally, we obtain 3970 documents (990 from rec.autos, 994 from
rec.motorcycles, 994 from rec.sport.baseball and 999 from rec.sport.hockey) and
a list of 8014 words. Each document will be treated as a point in a
8014-dimensional space.

To deal with text-kind data, we define a new distance introduced by Zhu et al.~\cite{ZhuGL03}: 
the distance between $x_{i}$ and $x_{j}$ is $d(x_{i},x_{j})=1-\cos{\alpha}$, where $\alpha$ 
is the angle between $x_{i}$ and $x_{j}$ in Euclidean space. Under this 
new distance, we ran the same experiment with the same parameter as we process the 
above MNIST dataset. 
The error bar of the tests for 20-newsgroups is presented in
Figure \ref{fig:er}~(b). A similar pattern result is observed, namely 
the PIM has the best performance when there are more than 2 labelled points in each class, 
and the GLM has the worst performance. 


\section{Conclusion}
We have presented two new approaches for solving the harmonic extension problem. 
Both are simple but have theoretical guarantees. We have also compared their performance
in the application of semi-supervised learning. In the future, we will test these 
methods on more datasets and find different applications of harmonic extension. 





\bibliographystyle{abbrv}
\bibliography{dirichlet}

\begin{thebibliography}{10}

\bibitem{BelkinN05}
M.~Belkin and P.~Niyogi.
\newblock Towards a theoretical foundation for laplacian-based manifold
  methods.
\newblock In {\em COLT}, pages 486--500, 2005.

\bibitem{CLEM_08}
M.~Belkin and P.~Niyogi.
\newblock Convergence of laplacian eigenmaps.
\newblock {\em preprint, short version NIPS 2008}, 2008.

\bibitem{mnist}
C.~J. Burges, Y.~LeCun, and C.~Cortes.
\newblock Mnist database.

\bibitem{Chung:1997}
F.~R.~K. Chung.
\newblock {\em Spectral Graph Theory}.
\newblock American Mathematical Society, 1997.

\bibitem{Doyle1984}
P.~G. Doyle and J.~L. Snell.
\newblock {\em Random Walks and Electric Networks}.
\newblock Mathematical Association of America, Washington, DC, 1984.

\bibitem{DuGLZ12}
Q.~Du, M.~Gunzburger, R.~B. Lehoucq, and K.~Zhou.
\newblock Analysis and approximation of nonlocal diffusion problems with volume
  constraints.
\newblock {\em SIAM Review}, 54(4):667--696, 2012.

\bibitem{Hein:2005:GMW}
M.~Hein, J.-Y. Audibert, and U.~von Luxburg.
\newblock From graphs to manifolds - weak and strong pointwise consistency of
  graph laplacians.
\newblock In {\em Proceedings of the 18th Annual Conference on Learning
  Theory}, COLT'05, pages 470--485, Berlin, Heidelberg, 2005. Springer-Verlag.

\bibitem{Lafon04diffusion}
S.~Lafon.
\newblock {\em Diffusion Maps and Geodesic Harmonics}.
\newblock PhD thesis, 2004.

\bibitem{LSS}
Z.~Li, Z.~Shi, and J.~Sun.
\newblock Finite integral method for solving poisson-type equations on
  manifolds from point clouds with convergence guarantees.
\newblock {\em arXiv:1409.2623}.

\bibitem{S15}
Z.~Shi.
\newblock Enforce the dirichlet boundary condition by volume constraint in
  point integral method.
\newblock {\em In preparation}.

\bibitem{SS14}
Z.~Shi and J.~Sun.
\newblock Convergence of the laplace-beltrami operator from point clouds.
\newblock {\em arXiv:1403.2141}.

\bibitem{Singer06}
A.~Singer.
\newblock {From graph to manifold Laplacian: The convergence rate}.
\newblock {\em Applied and Computational Harmonic Analysis}, 21(1):128--134,
  July 2006.

\bibitem{Singer13}
A.~Singer and H.~tieng Wu.
\newblock Spectral convergence of the connection laplacian from random samples.
\newblock {\em arXiv:1306.1587}.

\bibitem{ZhouBLWS03}
D.~Zhou, O.~Bousquet, T.~N. Lal, J.~Weston, and B.~Sch{\"{o}}lkopf.
\newblock Learning with local and global consistency.
\newblock In {\em Advances in Neural Information Processing Systems 16 [Neural
  Information Processing Systems, {NIPS} 2003, December 8-13, 2003, Vancouver
  and Whistler, British Columbia, Canada]}, pages 321--328, 2003.

\bibitem{Zhu:2005}
X.~Zhu.
\newblock {\em Semi-supervised Learning with Graphs}.
\newblock PhD thesis, Pittsburgh, PA, USA, 2005.
\newblock AAI3179046.

\bibitem{ZhuGL03}
X.~Zhu, Z.~Ghahramani, and J.~D. Lafferty.
\newblock Semi-supervised learning using gaussian fields and harmonic
  functions.
\newblock In {\em Machine Learning, Proceedings of the Twentieth International
  Conference {ICML} 2003), August 21-24, 2003, Washington, DC, {USA}}, pages
  912--919, 2003.

\end{thebibliography}




\end{document}